\pdfoutput=1

\documentclass[11pt]{article}

\usepackage{acl}

\usepackage{times}
\usepackage{latexsym}

\usepackage[T1]{fontenc}

\usepackage[utf8]{inputenc}

\usepackage{microtype}

\usepackage{microtype}
\usepackage{booktabs}
\usepackage{graphicx}
\usepackage{placeins}
\usepackage{enumitem}
\usepackage{soul}
\usepackage{url}
\usepackage{amsmath}
\usepackage{lipsum}  
\usepackage{tabularx}
\usepackage{tcolorbox}

\usepackage{todonotes}
\newcommand{\note}[4][]{\todo[author=#2,color=#3,size=\scriptsize,fancyline,caption={},#1]{#4}}
\newcommand{\aga}[2][]{\note[#1]{Aga}{red!20}{#2}}
\newcommand{\Aga}[2][]{\aga[inline,#1]{#2}\noindent}

\usepackage{xspace}

\newcommand{\kids}{K\xspace}
\newcommand{\teens}{T\xspace}
\newcommand{\women}{W\xspace}
\newcommand{\men}{M\xspace}

\newcommand{\guidetitle}[1]{``#1''\xspace}
\newcommand{\exampleword}[1]{`#1'\xspace}

\newcommand{\wmGr}{W--M\xspace}
\newcommand{\ktGr}{K--T\xspace}
\newcommand{\ourdata}{\textcolor{black}{wikiHowAudiences}\xspace}

\newcommand{\body}{\textsc{Body}\xspace}
\newcommand{\interactions}{\textsc{Interact}\xspace}
\newcommand{\presentations}{\textsc{Present}\xspace}
\newcommand{\grownups}{\textsc{Grown-up}\xspace}
\newcommand{\advices}{\textsc{Advice}\xspace}
\newcommand{\activities}{\textsc{Activity}\xspace}

\usepackage{cleveref}
\usepackage{multirow}

\title{How-to Guides for Specific Audiences: A Corpus and Initial Findings}

\author{Nicola Fanton$^{1}$ \\[-0.2em]
  \footnotesize{he/they} \\
  \\\And
  Agnieszka Falenska$^{2}$ \\
  University of Stuttgart \\
  $^{1}$ Institute for Natural Language Processing \\
  $^{2}$ Interchange Forum for Reflecting on Intelligent Systems \\
  \texttt{\{firstname.lastname\}@ims.uni-stuttgart.de} \\
  \\\And
  Michael Roth$^{1}$ \\
  }

\begin{document}
\maketitle
\begin{abstract} 
Instructional texts for specific target groups should ideally take into account the prior knowledge and needs of the readers in order to guide them efficiently to their desired goals. However, targeting specific groups also carries the risk of reflecting disparate social norms and subtle stereotypes. In this paper, we investigate the extent to which how-to guides from one particular platform, wikiHow, differ in practice depending on the intended audience. We conduct two case studies in which we examine qualitative features of texts written for specific audiences. In a generalization study, we investigate which differences can also be systematically demonstrated using computational methods. The results of our studies show that guides from wikiHow, like other text genres, are subject to subtle biases. We aim to raise awareness of these inequalities as a first step to addressing them in future work. 
\end{abstract}

\section{Introduction}\label{sec:introduction}

\emph{How-to guides} provide practical instructions that help humans to achieve specific goals. In the past decades, such guides also attracted increasing interest in NLP and AI research \cite{branavan-etal-2009-reinforcement,chu2017distilling,anthonio-etal-2020-wikihowtoimprove}. 
Resources such as wikiHow,\footnote{\url{www.wikihow.com}} a collaboratively edited online platform for instructional texts, make it possible to scale research efforts to hundreds of thousands of articles.\ 
By covering an ever-increasing number of guides, including niche topics and articles for minority groups, there is also an increasing risk of perpetuating stereotypes and jeopardizing general accessibility. In fact, we notice that wikiHow already contains articles written for specific target groups as well as articles that exist in different versions for different audiences. As an example, Table~\ref{tab:act-like-a-kid-again} shows two articles with the same \emph{title}, \guidetitle{Act Like a Kid Again}, one with the \emph{indicator} \exampleword{(Girls)} and one with \exampleword{(Boys)}.

\begin{table}[t]
\centering
\begin{tabularx}{.45\textwidth}{p{.9\linewidth}}
    \toprule
    \multicolumn{1}{c}{Act Like a Kid Again ({Girls})} \\
    \textbf{Eat well and exercise}, but don’t obsess about your body. Be healthy without stressing too much about it. (\ldots)
    Generally, \textbf{go for lots of fruits and veggies}. And even though kids love sugar, \textbf{don’t eat too much} of it!
    \\
    \midrule
    \multicolumn{1}{c}{Act Like a Kid Again ({Boys})} \\
    Eat your childhood favorite food. Recollect \textbf{every snack}, chocolates, ice cream, candy bars, cotton candy and \textbf{everything that you loved as a kid} or would make you feel pampered. Eat as per your capacity as too much at once may make you feel uncomfortable.
    \\
    \bottomrule
\end{tabularx}
    \caption{Two versions of the same guide in wikiHow.}
    \label{tab:act-like-a-kid-again}
\end{table}

Among other things, we find that such articles dramatically differ in terms of details. For example, the texts highlighted in Table~\ref{tab:act-like-a-kid-again} vary in how much they focus on issues potentially related to body images. As such, the articles reflect \textit{disparate standards}, which ultimately may contribute to discrimination \cite{prentice2002women}. The specific example can also be linked to observations of gender differences in weight concerns from psychology \cite{dougherty2022stress}, which might represent a reason for \textit{disparate treatment}. 
On the surface, it is not always possible to say exactly why there are certain differences in articles for specific audiences. However, through qualitative and quantitative comparisons on the linguistic level, we can at least determine what types of differences are present and to what extent they can be systematically identified. In this sense, we aim to contribute to questions about biases and fairness in data and, at the same time, connect to related research in psychology and other social sciences.

There already exists a large body of research that examines biases and stereotypes in NLP data and, 
likewise, how-to guides from wikiHow have been used as 
training material for a variety of language processing tasks
(\S\ref{sec:related-work}). However, previous studies have not explicitly looked into issues related to bias in the wikiHow data. 
As a first step towards addressing this gap, we create our own sub-corpora of how-to guides, which let us investigate differences across articles for specific target groups~(\S\ref{sec:corpus}).

We perform two case studies and a generalization study on our collected data:
In the first study, we identify a number of articles that exist in multiple variants for different target groups and examine them in terms of distinctive content and linguistic characteristics (\S\ref{sec:case-study}). 
As a second case study, we explicitly examine how far topics covered for specific target groups differ from each other~(\S\ref{sec:howtobe}).
Finally, we investigate whether the qualitative findings from our case studies can be validated quantitatively and generalized to our whole corpus using computational modeling (\S\ref{sec:modeling}). 

In summary, we find systematic differences between articles for specific groups in terms of topic, style, and content. We conclude the paper with a discussion of these findings and point out links to existing work in the social sciences (\S\ref{sec:discussion}).

\section{Related Work}\label{sec:related-work}

We summarize existing work on the three strains of research that this paper builds on: wikiHow as a data source (\S\ref{sec:instructions}), subtle biases in datasets (\S\ref{sec:bias-nlp}),
as well as understanding the characteristics of texts that target specific audiences (\S\ref{sec:different-audiences}).

\subsection{wikiHow as a Data Source}\label{sec:instructions}
wikiHow is a prominent data source for a variety of tasks, including summarization \cite{koupaee2018}, goal-step inference \cite{zhang-etal-2020-reasoning}, and question answering \cite{cai-etal-2022-generating}. 
By exploiting the revision history of wikiHow, \citet{anthonio-etal-2020-wikihowtoimprove} created 
\textbf{wikiHowToImprove}, which has been used to better understand phenomena related to the (re-)writing process of how-to guides \citep{roth-anthonio-2021-unimplicit,anthonio-etal-2022-clarifying}. 
Writing, but especially revising, instructions should presumedly take into account the readers' context, perspective and knowledge about the domain and the world. 
The need for clarification stands prominently out as a main purpose of the refinements of wikiHow guides \cite{bhat-etal-2020-towards}. It has been shown that while annotators tend to agree that ``revised means better'', the disagreements can be caused by differences in common knowledge and 
intuitions \cite{anthonio-roth-2020-learn}. As specific phenomena, previous work studied implicit references and lexical vagueness \cite{anthonio-roth-2021-resolving, debnath-roth-2021-computational}. However, none of the aforementioned studies accounted for audience-specific differences. This work takes a first step to close this gap.

\subsection{Subtle Biases in Datasets}\label{sec:bias-nlp}
Diagnosing the presence of biases in data is one of the crucial steps in diminishing the spread of harmful stereotypes.
This work contributes to the research on \emph{subtle biases}, i.e.,  textual patterns that implicitly reflect societal power asymmetries.
Such biases are embeded in specific linguistic phenomena (e.g., masculine generics; \citeauthor{swim-etal:2004:SR}, \citeyear{swim-etal:2004:SR}) or in inequalities in how people from different demographic groups are represented (e.g., emphasizing the romantic relationships in the bibliographies of women; \citeauthor{wagner2015}, \citeyear{wagner2015}). 
Moreover, they can be frequent even in domains where blatant stereotypes and openly expressing beliefs about social hierarchies is generally considered inappropriate \citep{cervone-etal:2021:LSP}. 
For example, there is a long line of work analyzing subtle stereotypes in Wikipedia \citep[among others]{callahan2011cultural,reagle-rhue:2011,konieczny2-klein:2018,schmahl-etal:2020:workshop_css}, where the lack of diversity represents an issue already at the level of the editors' community \cite{lam2011}. 
Beyond notability for representation itself, linguistic aspects in Wikipedia show a remarkable disparity concerning biographies of men and women, both in terms of topics and polarity of abstract terminology \cite{wagner2016}.
Such inequalities do not pertain only to biographies but find systemic correspondence in all domains and across languages \cite{falenska-cetinoglu-2021-assessing}. 

To the best of our knowledge, the presence of subtle stereotypes in wikiHow has not yet been investigated. However, the guides from this platform are a valuable entry point for studying 
bias, as they are produced by a community of contributors and by experts\footnote{\url{https://www.wikihow.com/Experts}} suggesting how to perform activities. In other words, given the different purposes of the platforms, while Wikipedia data is rather descriptive, wikiHow data features instructional texts that potentially differ depending on the audience. 

\subsection{Different Audiences}\label{sec:different-audiences}
The mind of the readers features a priori goals that affect the understanding of written texts \cite{fum-etal-1986-tailoring}. 
However, 
the goals and knowledge of different (groups of) people may vary. 
An example of work that considers different readers' expertise regards title generation \cite{senda-shinohara-2002-analysis}. 
In that work, less expert readers were found to be tentatively more influenced by effective titles. 
Consequently, a system for revising titles accounting for the readers' expertise has been proposed \cite{senda-etal-2004-support}. As such, that contribution indicates the importance of considering the target audience for efficient communication. Additionally, different audiences can understand to different extents technical terminology \cite{senda-etal-2006-automatic, elhadad-sutaria-2007-mining} and causation \cite{siddharthan-katsos-2010-reformulating}. Previous contributions accounted for different target groups also in the controllable text generation tasks of paraphrasing \cite{kajiwara-etal-2013-selecting}, 
text simplification \cite{scarton-specia-2018-learning, sheang-saggion-2021-controllable}, machine translation \cite{agrawal-carpuat-2019-controlling}, and dictionary examples generation \cite{he-yiu-2022-controllable}.

\section{Corpus Construction}\label{sec:corpus}


As introduced in \S\ref{sec:instructions}, wikiHowToImprove is a well-established data set derived from wikiHow and consisting of 
more than 246,000 how-to guides. In general, each guide consists of multiple revisions of an \textit{article}, a fixed goal that is named in the \textit{title}, and (optionally) an \textit{indicator} that follows the title in parentheses (cf.~Table~\ref{tab:act-like-a-kid-again}).
%
As we are interested in how-to guides for different target groups, we filter the data for indicators that specify a group of people as targets, which we also refer to as the \textit{audience}. Table~\ref{tab:distrib-articles} lists the 20 most frequent indicators extracted from wikiHowToImprove.

\begin{table}[t]
    \centering
    \begin{tabular}{@{}rlr|rlr@{}}
    \toprule
    \multicolumn{3}{@{}l}{\textbf{Rank} \quad \textbf{Indicator} \hfill \textbf{\#}} &
    \multicolumn{3}{l@{}}{\textbf{Rank} \quad \textbf{Indicator} \hfill \textbf{\#}} \\
    \midrule
    1 & Girls & 370 & 11 & Guys & 35 \\
    2 & for Girls & 284 & 12 & for Women & 35 \\
    3 & for Kids & 182 & 13 & Women & 34 \\
    4 & Kids & 114 & 14 & \underline{UK} & 34 \\
    5 & Teens & 110 & 15 & for Men & 31 \\
    6 & Teen Girls & 100 &  16 & \underline{Christianity} & 31 \\
    7 & for Teens & 73 & 17 & Men & 29 \\
    8 & \underline{USA} & 49 & 18 & \underline{for Beginners} & 29 \\
    9 & for Guys & 42 & 19 & Boys & 25 \\ 
    10 & \underline{Windows} & 38 & 20 & Teenage Girls & 25 \\ 
    \bottomrule
    \end{tabular}
    \caption{Counts of the 20 most frequent indicators.}
    \label{tab:distrib-articles}
\end{table}

Based on a manual grouping of these indicators, we find that 15 out of 20 indicators refer to attributes of performative gender and age (the remaining five are underlined in \Cref{tab:distrib-articles}). Apart from their high frequency, both of these attributes are of interest to studies in the social sciences, in which they are often used as independent variables \cite{cortina2013selective,cha2014overwork,palencia2014influence}.
Following a traditional binary setup, we distinguish two audiences based on gender, women (\textbf{\women}) and men (\textbf{\men}), and two audiences based on age, kids (\textbf{\kids}) and teens (\textbf{\teens}).\footnote{Note that while the selected audiences follow discrete categories, we explicitly caution that individual readers can only be represented on a continuum.} For each type of audience, we create a set of all indicators used and collect all corresponding guides by extracting the latest article versions from wikiHowToImprove.

\begin{table}[t]
    \centering
        \begin{tabular}{lrr|rr}
         \toprule
        \quad \textbf{} & \multicolumn{1}{c}{\textbf{W}} & \multicolumn{1}{c}{\textbf{M}} & \multicolumn{1}{c}{\textbf{K}} & \multicolumn{1}{c}{\textbf{T}} \\
        \midrule
        Indicators & 29 & 13 & 23 & 16 \\
        Articles & 993 & 209 & 499 & 411 \\
        Sentences per article & 40 & 50 & 29 & 43  \\
        Words per article & 509 & 682 & 352 & 544  \\
        \bottomrule
        \end{tabular}
    \caption{The distribution of the indicators and of the articles for the target audience groups. Sentences and words are indicated via their median values by article.}
    \label{tab:distribution_}
\end{table}

Statistics of our corpus with audience-specific how-to guides are provided in Table~\ref{tab:distribution_}. We note that there is a much higher number of indicators and articles for \women than for \men. In comparison, the number of articles and indicators for \kids and \teens are similar. With only 2,112 how-to guides in total, the corpus seems relatively small. However, the average length of articles ranges
from 352 to 682 words, which adds up to a corpus size of more than one million words. Throughout this work, we refer to this dataset as \ourdata.\footnote{\url{https://github.com/mnfanton/wikiHowAudiences}}
Next, we approach it in its entirety with two case studies.




\section{Case Study: Same Title, Different Audience}
\label{sec:case-study}

\begin{table}[t]
\centering
\begin{tabular}{@{}lll@{}}
    \toprule
    \multicolumn{3}{c}{\textbf{Women -- Men}} \\
    \midrule
    \body & 11 & Lose Belly Fat \\
    \interactions & 11 & Act on a Date  \\
    \presentations & 13 & Dress Like a CEO \\
    \midrule
    \multicolumn{3}{c}{\textbf{Kids -- Teens}} \\
    \midrule
    \grownups & 3 & Look Older \\
    \advices & 4 & Balance School and Life \\
    \activities & 10 & Apply Makeup \\
    \bottomrule
\end{tabular}
    \caption{Frequencies and examples of topical categories.}
    \label{tab:same_title}
\end{table}

Our starting example from \Cref{tab:act-like-a-kid-again} includes two guides with the same title but different target indicators. Such guides outline the ultimate instances of instructions that are written for different audiences. Therefore, we start our investigation by analyzing how often such cases occur in \ourdata, which topics they cover, and what differs between versions for specific target groups. 

\subsection{Guides Selection}
First, we identify titles that occur more than once in \ourdata: 32 unique titles for \wmGr and 15 for \ktGr. Next, we group guides with the same title but different target audiences into pairs. 
A complete list of article titles in this subset can be found in \Cref{appendix:case-study}. 


\subsection{Guides Analysis}
To understand which goals require audience-specific adaptations, we analyze the topics and articles of the filtered guides.


\paragraph{Topics.} 
We start by manually investigating titles of the filtered pairs of guides. For this purpose, we assign each of them to one of three content-related categories. The categories were designed to cover all the titles while being as concrete as possible. An overview of all the categories and their examples is listed in \Cref{tab:same_title}.

We find that \wmGr instructions cover a relatively wide range of topics, from body-related activities (\body), over interacting with other people (\interactions), to self-presentation (\presentations), which is the most frequent category. 
In contrast, among titles in \ktGr, we notice one clear pattern: all topics focus on issues that require different steps depending on the age of the target. Among them, we distinguish and report in ascending order of frequency articles about learning how to do activities for grown-ups or concerning the urge to grow old (\grownups), advice related to the life of young people (\advices), and activities about oneself or the relation of oneself to others (\activities).


\paragraph{Length.} Next, we check whether there are significant differences 
in terms of how detailed the instructions are for different target groups. We quantify this by simply measuring the length per article in words and sentences. 
We notice a considerable difference between \kids and \teens: the median length of articles for \kids is only 30 sentences and 346 words, while articles for \teens contain 98 sentences and 1081 words. In the case of \women and \men, we do not find such large differences in terms of average word (785 vs.~856) and sentence counts (59 vs.~62). Overall, the numbers reflect the patterns shown in \Cref{tab:distribution_} for the whole \ourdata data. 

\paragraph{Content.} Finally, we switch our attention to the actual content of the articles. As a simple measure of how similar two guides are, we consider their word overlap in both directions using BLEU score \cite{papineni-etal-2002-bleu-custom}. 

\begin{table*}[t]
\centering
\begin{tabular}{@{}cp{6.8cm}l@{~}p{6.8cm}@{}}
    \toprule
   \wmGr & \multicolumn{1}{c}{Get Clear Skin (0.02 BLEU)} && \multicolumn{1}{@{~}c@{~}}{Recognize Chlamydia Symptom (0.69 BLEU)} \\
    \cmidrule{2-2} \cmidrule{4-4}
\multirow{3}{0.3cm}{\women} & Gently pat your face dry with a clean towel. Don’t rub your face! This can irritate your skin more. && Chlamydia is a dangerous yet common and curable sexually transmitted infection (...) \\
\multirow{3}{0.3cm}{\men} & Dry your face -- but not roughly. && Chlamydia, specifically chlamydia trachomatis, is a common and curable but dangerous sexually transmitted infection (...) \\
\midrule
\ktGr & \multicolumn{1}{c}{Flirt (0.05 BLEU)} && \multicolumn{1}{c}{Make Money (0.59 BLEU)} \\
    \cmidrule{2-2} \cmidrule{4-4}
\multirow{3}{0.3cm}{\kids} & Make eye contact. Both girls and boys love eye contact. && There are the traditional jobs like babysitting, shoveling snow, and doing chores around the house. \\
\multirow{3}{0.3cm}{\teens} &  Make eye contact. Body language is a big part of flirting, and a big part of that is eye contact. Eye contact conveys intimacy (...) && Babysit for friends and family. One of the best ways for teenagers to make money and help out in the community is babysitting. \\

\bottomrule
\end{tabular}
    \caption{Excerpts from the article pairs with the lowest (left) and highest (right) word overlap.}
    \label{tab:word_overlap}
\end{table*}

\Cref{tab:word_overlap} presents the articles with the lowest and highest word overlap in both analyzed groups. Interestingly in the case of \wmGr, both articles cover concepts related to \body, namely clearing skin and recognizing an infection. Manual inspection of their content reveals that even in the case of the least overlapping articles, \guidetitle{Get Clear Skin}, slight differences can be noticed: \women article includes more specific information as well as different usage of punctuation. In the case of most overlapping articles, \guidetitle{Recognize Chlamydia Symptoms}, the main difference comes from the vocabulary related to different body parts from body types. 
The high word overlap of these two versions is likely related to their introductions, which provide an interchangeable overview to the topic. 

In the case of \ktGr, the least and most overlapping articles come from two different categories: \activities and \grownups. The least overlapping pair, \guidetitle{Flirt}, is a case of two instructions that treat the same goal with different levels of complexity. For example, the matter of eye contact is described with one step in \kids and more than ten in \teens. 
The most overlapping articles, \guidetitle{Make Money}, can be an example of a content stalemate -- 
for both target audiences, babysitting is the first suggested activity to achieve the profit goal. 
However, it is possible to notice differences in how this concept is contextualized for two groups: either in a list of activities or discussed with its implications and advantages.

\subsection{Summary} 


We exemplified three characteristics that can distinguish guides written for different audiences. First, the instructions written for \ktGr significantly differed in \emph{length}. Next, we saw pairs of guides that varied in \emph{style} (such as punctuation) and \emph{content} (e.g., vocabulary in \body articles). 
Some of the presented examples suggest that considering only simple content features could be enough to distinguish articles written for different audiences. However, such an approach could be insufficient in more complex cases, such as pairs of guides with high word overlap (see \guidetitle{Make Money}).
We discuss these articles again in our generalization study (\S\ref{sec:modeling}).

\section{Case Study: ``How To Be'' Guides}\label{sec:howtobe}

In the previous section, we looked at how-to guides that occur in different versions
for specific audiences. Such guides might concern particular goals that \emph{require} being addressed in distinct ways. In this section, in contrast, we broaden the scope of analysis to explore other cases of differences in audience-specific instructions. 

\subsection{Guides Selection}

The initial example from the introduction (see \Cref{tab:act-like-a-kid-again}) explain how to perform like somebody the reader is presumably not. Inspired by this example, we investigate what other guides instruct their readers ``how to be''. Concretely, we filter titles starting with the word `be', which gives us 118 guides for \women, 20 for \men, 32 for \kids, and 30 for \teens. 

\subsection{Guides Analysis}


To understand which topics the ``how to be'' guides cover, we group them according to the first word that occurs after 
\exampleword{be} (henceforth the \emph{completion}).\footnote{We ignore the articles `a', `an', and `the'.} \Cref{tab:how-to-be-examples} shows the most frequent completions for each target group and respective example titles. 


Regarding \ktGr guides, we notice no clear pattern that would distinguish instructions based only on their titles. There is roughly the same number of how-to articles for \kids and \teens (32 vs.\ 30). Moreover, among the most frequent completions we commonly find the word `good', followed by words such as `comfortable', `less', or `safe'.

\begin{table}[t]
    \centering
    \begin{tabular}{@{}lll@{}}
    \toprule
    \multicolumn{2}{l}{\textbf{Completion(s)}} & \textbf{Title} \\
    \midrule
     \multirow{2}{*}{\women} & Popular & Be \underline{Popular} and Athletic \\
     & Cute & Be \underline{Cute} at School \\
     \midrule
     \multirow{2}{*}{\men} & Cool & Be \underline{Cool} in High School \\
     & More & Be \underline{More} Physically Attractive \\
     \midrule
     \kids & Good & Be \underline{Good} With Money \\
     \midrule
     \teens & Good & Be a \underline{Good} Friend \\
     \bottomrule
    \end{tabular}
    \caption{The most frequent target-specific completions of "how to be" guides and examples of respective titles. 
    }
    \label{tab:how-to-be-examples}
\end{table}

In contrast, we find substantial differences for \wmGr. Specifically, we note that ``how to be'' guides are more common for \women (12\% of all articles for this target group)
and for both audiences we find differing frequencies of completions:
While \women articles focus on being `cute' and `popular' (9 guides), \men articles put more emphasis on being `cool' and `more' (6 guides). Even though all the how-to guides refer to similar contexts (mostly related to school), we do not find mutual correspondence---there are no instructions for how to ``be cool at school'' for \women and no guide for how to ``be cute at school'' for \men.

\subsection{Summary}
In this section, we looked at a particular subset of \ourdata, namely guides with titles starting with the word `be'. We found that, in the case of \wmGr targets, the differences in instructions occur already at the level of goals that these guides describe. In other words, we saw examples of instructions where the information for which audience they were intended could be deduced strictly from their \emph{titles}.

\section{Generalization Study: Computational Approach}\label{sec:modeling}

Our case studies show that, depending on the audience, there exist examples of articles that differ in terms of topic, length, style, and/or vocabulary. 
However, an open question is whether these are only individual cases or if such differences occur systematically. In this study, we investigate this question computationally and attempt to verify our observations on the basis of a larger dataset. For this purpose, we implement tentative characteristics in the form of features and models (\S\ref{sec:features}), evaluate in a setting with our full sub-corpora (\S\ref{sec:setup}), discuss quantitative results (\S\ref{sec:classifiers}), and analyze qualitative findings (\S\ref{sec:error-analysis}).


\subsection{Models}\label{sec:features}

Based on the findings from the two case studies, we define majority and length-based baselines and several simple logistic regression classifiers with different sets of features.


\paragraph{Baselines.} We use a simple majority baseline that always assigns the most frequent class. We also implement two length-based baseline models that use the number of words in a title (or article) as the only feature for classification.

\paragraph{Content (title/article).} The words and phrases used in a text can be potential indicators of its target group. Thus, we make use of the most common\footnote{Note that we could have used all n-grams, but due to the small size of our data (see~\S\ref{sec:setup}), we decided to limit the number of features via an additional hyperparameter.} uni-grams and bi-grams, excluding stop words, as a feature representation for the content of a how-to guide. We evaluate two variants: features derived from the articles and from the titles. 
\paragraph{Style (article).} We represent style using two sets of established features from authorship attribution \cite{sari-etal-2018-topic}, namely \textit{lexical} style: average word length, number of short words, vocabulary richness in terms of hapax-legomena and dis-legomena, \% of digits, \% of upper case letters; and \textit{syntactical} style: occurrences of punctuation, frequencies of POS tags, and stop-word frequencies. 

\paragraph{\texttt{combined} (article).} Content and style can potentially provide complementary information. We test whether a model can leverage a combination of information from different sources. For this purpose, we simply concatenate the article-level features for content, style, and length.

\paragraph{RoBERTa (article).} As an alternative to manually selected features, we further test features derived from a large language model, RoBERTa \cite{roberta}. Specifically, we encode the article's text, truncated to the first 512 tokens, and extract the representation of the 
special classification token
from the last hidden layer as a set of feature values. 

\subsection{Experimental Setup}
\label{sec:setup}

\begin{table}[t]
    \centering
        \begin{tabular}{lrr|rr|rr}
         \toprule
        \quad \textbf{} & \multicolumn{1}{c}{\textbf{W}} & \multicolumn{1}{c}{\textbf{M}} & \multicolumn{1}{c}{\textbf{K}} & \multicolumn{1}{c}{\textbf{T}} & \textbf{Total} \\
        \midrule
        \textsc{Train}  & 805 & 172 & 416 & 337 & 1,730\\
        \textsc{Dev}  & 94 & 23 & 45 & 37 & 199 \\
        \textsc{Test}  & 94 & 14 & 38 & 37 & 183 \\
        \midrule
        Total & \multicolumn{2}{c}{1,202} & \multicolumn{2}{c}{910} & 2,112\\
        \bottomrule
        \end{tabular}
    \caption{Number of articles for each target group and data split, as well as for each task in total.}
    \label{tab:distribution_2}
\end{table}

In order to find out whether and to what extent articles for different target groups can be distinguished computationally, we define two classification tasks in which specific articles, based on their characteristics, are to be assigned to one target group each. We distinguish between articles for women and men (\wmGr) and between articles for kids and teenagers (\ktGr). For all four classes, we use the full \ourdata, which we divide into \textsc{train}, \textsc{dev}, and \textsc{test} sets following the article-level partition of the original wikiHowToImprove corpus \cite{anthonio-etal-2020-wikihowtoimprove}. Statistics for each class and set are shown in Table~\ref{tab:distribution_2}.
For the style features, the texts are lemmatized with spaCy.\footnote{\texttt{https://spacy.io/}}

We train each model on the \textsc{train} set and evaluate in terms of macro F$_{1}$-score on the \textsc{test} set. We compute F$_{1}$-score per class as the harmonic mean between precision (ratio of correct predictions) and recall (ratio of correctly classified instances). As our data is imbalanced, we use macro F$_{1}$ instead of a weighted/micro score to treat each class (rather than each instance) as equally important. 

A number of hyperparameters are optimized on the \textsc{dev} set: We try different values for the logistic regression classifiers' L1 and C terms, sampled from 10 instances between $1e-5$ and $100$. 
For the content features, we optimize the number of $k$ most 
common n-grams ($k=200$). We also made use of the \textsc{dev} set to determine the best language model for our tasks, which we found to be \texttt{roberta-large} (results of other models are shown in Appendix~\ref{sec:appedix-classification}).\footnote{We used HuggingFace Transformers \cite{wolf-etal-2020-transformers}.}

\subsection{Results}\label{sec:classifiers}

\begin{table}[t]
    \centering
        \begin{tabular}{lr|r}
        \toprule
        \textbf{Model} & \multicolumn{1}{c}{\textbf{W--M}} & \multicolumn{1}{c}{\textbf{K--T}} \\
        \midrule
        \multicolumn{3}{l}{\quad \textbf{Baselines}} \\
         Majority baseline & 0.47 & 0.34 \\
        Length (title) & 0.47 & 0.44 \\
        Length (article) & 0.47 & 0.61 \\
        \cmidrule{1-1}
        \multicolumn{3}{l}{\quad \textbf{Content \& Style}} \\
        Content (title) & 0.57 & 0.57 \\
        Content (article) & 0.59 & 0.78 \\
        Style (article) & 0.58 & 0.67 \\
        \cmidrule{1-1}
        \multicolumn{3}{l}{\quad \textbf{``Full'' models}} \\
        \texttt{combined} (article) & 0.71 & 0.78\\
        RoBERTa (article) & 0.68 & 0.74 \\
        \bottomrule
        \end{tabular}
    \caption{Macro F$_{1}$-scores on the test sets.}
    \label{tab:classification_}
\end{table}

The results are summarized in Table~\ref{tab:classification_}. As conjectured based on the \ktGr articles from the first case study, we find that the length-based baselines indeed outperform the majority baseline\footnote{Note that the F$_1$-score for the majority baseline lies below 0.5 because we calculate the \textit{macro average} over both classes and the score does not reach 1.0 for either class.} in that setting. As the further results show, content and stylistic features can indeed be used to correctly assign a specified target group to many how-to guides. According to the evaluation scores, features calculated at the article level are particularly suitable for this purpose:
The \texttt{combined} model, which uses content, style and length features on the article level, achieves the best result with macro-F1 scores 
of 0.71 and 0.78 for \wmGr and \ktGr, respectively.
Features generated based on the \texttt{roberta-large} language model achieve competitive scores (0.68 and 0.74), but fall short of the 
\texttt{combined} 
model.

The large differences in result between the baselines and our models show that the target audience of many articles can be determined simply from the vocabulary and style of an article. Next, we take a closer look at model features and errors.


\subsection{Analyses}\label{sec:error-analysis}

For our analyses, we focus on the \texttt{combined} model because it achieves the best results and its features are easily interpretable. 

\begin{table*}[t]
  \centering
  \begin{tabular}{rp{2cm}p{6.3cm}p{5.5cm}}
    \toprule
    \textbf{} & \textbf{Feature(s)} & \textbf{Example} & \textbf{Title} \\
    \midrule
    \textbf{W} & cute, makeup & Do \ul{cute} \ul{makeup}. & Look Cute \\ 
       & wasn't & She most likely \ul{wasn't} wearing the right colors for her skin tone. & Go from Ugly to Popular \\
    \midrule
    \textbf{M} & hers & Slowly move your hand towards \ul{hers} \ldots & Know if Your Crush Likes You Back \\
      & theirs & Being a good partner is all about \ldots adjusting your style to suit \ul{theirs}. & Grind \\
    \midrule
    \textbf{K} & name & Think of your blog's \ul{name}. &  Write a Blog  \\
      & kid & \ldots even if you’re a \ul{kid}, there are ways to bank a few extra bucks. & Make Money \\
    \midrule
    \textbf{T} & dress & \ul{Dress} up, make it look important. & Know What to Wear on Dates \\
      & teen & When you’re a \ul{teen} with a busy schedule, it can be difficult to find time to be active. & Stay Active After School \\
    \bottomrule
    \end{tabular}
    \caption{Sample of the top-10 most predictive features and example sentences from articles of each target group.
    } 
    \label{tab:bigtable}
\end{table*}

\paragraph{Features.} For each target group, we analyze what features are most important to the model. Since our model uses independent features in a binary classification task, we can simply check the highest positive and negative feature weights for this purpose. A selection from the ten most predictive features\footnote{Appendix~\ref{sec:appedix-classification} lists all top-10 most predictive features.} and example sentences are shown in Table \ref{tab:bigtable}. As the examples illustrate, some of the strongest features are, again, based on stereotypes (e.g., \exampleword{cute}, \exampleword{makeup} for \women) or reflect heteronormative assumptions (\exampleword{hers} for \men). Interestingly, we also see characteristics of gender-inclusive language (\exampleword{theirs} for \men) and direct address of the reader in terms of their group membership (\exampleword{kid} for \kids and \exampleword{teen} for \teens). We further find negations (e.g., \exampleword{wasn't}) as part of strong features for \women, which is particularly worrying in light of sociopsychological findings that have shown negations to serve a stereotype-maintaining function across languages \cite{Beukeboom2010,beukeboom2020negationbias}.


\paragraph{Same title articles.} 
As examples of particularly hard cases, we return to the how-to guides from the first case study, which consisted of article versions for different audiences~(\S\ref{sec:case-study}). Following the data partition from previous work, we identify 16 such articles in the \textsc{dev} and \textsc{test} splits. We find that the \texttt{combined} model classifies 12 of them correctly (75\%). In the remaining 4 cases, the prediction errors could have been caused by superficial features that are predictive for the opposite audience.
We note for each of these 16 articles that the version for the opposite audience is part of the \textsc{train} split. Therefore, the topics of the guides are generally not specific to one audience, and a correct classification of the majority of cases demonstrates that the model indeed captures characteristics of content and style that seem specific to the audience itself.

\section{Discussion and Conclusion}\label{sec:discussion}

In this paper, we assessed differences across how-to guides written for specific audiences. In the construction of sub-corpora for four target groups, we already noticed inequalities on the level of who is being instructed in wikiHow: as a target audience, women are mentioned more than four times more frequently than men, and teens receive about 50\% more instructions per article than kids. In two case studies, we investigated and provided examples of target-related differences on the levels of topic, style, and content. 

The differences observed in our case studies inspired feature sets of shallow classifiers for predicting the target audience of a given guide. Using these classifiers, we showed that it is, in many cases, indeed possible to automatically predict for which audience an article was written. In an analysis of our results, we found that this success is not merely based on different topics covered for each target group but that the articles for each group systematically differ in terms of content and style. 

Each of the aforementioned observations presents a tiny, seemingly insignificant piece of a puzzle. But taken together, these pieces reveal a surprisingly clear picture: there are noticeable differences in what topics are covered for each target group, how many articles and instructions are provided for each audience, and how these articles are written.
Even though the audience-specific characteristics 
used in our studies are by no means exhaustive, our straightforward approach allowed us to 
identify, qualitatively and quantitatively, debatable differences in how wikiHow guides present particular topics to specific target groups. 
While there is an inevitable need for differences in vocabulary when speaking about physical features or body parts, 
it is at best unclear in which ways how-to guides about human interactions or self-presentation should cast significant differences.

Some of the observed differences have already been critically discussed in the context of social science research. 
For example, it is well-known that labels such as `cute' are used pejoratively as a form of social control \cite{talbot2019language} and that prescriptive components of gender stereotypes in education contribute to discrimination \cite{kollmayer2018gender}. 
However, exposing readers to cultural messages and beliefs about age, gender or other factors cannot be avoided entirely, especially on a collaboratively edited online platform. In fact, it seems to be a challenge for any pluralistic society to find a balance between communicating traditional values and empowering everyone.
It is therefore all the more important for a comprehensive understanding to determine when and in what form social norms are conveyed.
As such, we view the contributions of this paper, 
namely our data set of audience-specific guides, 
\ourdata, %
and 
our mixed-methods approach for identifying and verifying differences, 
as a valuable connecting point to raise awareness of potential issues and to foster interdisciplinary dialogue for future research.

\section*{Limitations}\label{sec:limitations}
Our studies focus on the differences in how-to 
guides written for specific audiences only in one language, namely English. 
A major limitation is therefore that we do not consider other languages.

The perspectives provided by the data source we rely on, wikiHow, allow us to identify specific phenomena and peculiarities. Yet, contemplating only one data source lets us generalize only to a limited extent.
For example, the audiences considered in this work depended on the target groups portrayed in the data. They are neither exhaustive nor representative of the diversity of humankind, especially of marginalized social groups.
Therefore, a wider variety of data sources will be needed to test generalizations. 

Finally, a further limitation of our studies concerns intersectionality. While it seems possible that guides can be tuned by contemplating one specific attribute of the audience at a time, this does not hold with regard to the actual attributes of the readers. Such attributes are per se coexistent, and consequently, they are not separable.

\section*{Ethics Statement}\label{sec:ethics}
We acknowledge that the content that emerged from the data is narrow in terms of cultural perspectives, mainly addressing western cultures. 
Moreover, the analysis of the audiences is not exhaustive of the diversity of humankind, especially not exhaustively accounting for queer identities in particular trans and non-binary identities. 
With the present research, we do not intend to reinforce representational biases, rather to highlight them.

\bibliography{anthology,custom}
\bibliographystyle{acl_natbib}

\clearpage
\appendix
\section{Appendix}
\label{sec:appendix}

\subsection{Case Study}\label{appendix:case-study}

\begin{table}[!ht]
    \centering
    \small
    \begin{tabular}{rrlll}
    \toprule
    Category & \begin{tabular}[c]{@{}r@{}}Word\\Overlap \end{tabular}& Same Title & Indicator for \textbf{W} & Indicator for \textbf{M} \\
    \midrule
    \textsc{Body} & 0.02 & Get Clear Skin & for Middle School Girls & Guys \\
    \textsc{Body} & 0.05 & Burn Fat & for Girls & for Men \\
    \textsc{Present} & 0.07 & Get Ready for School & for Girls & Guys \\
    \textsc{Present} & 0.09 & Look Rich Without Being Rich & Teen Girls & for Guys \\
    \textsc{Present} & 0.16 & Get Ready for School & Teen Girls & Guys \\
    \textsc{Interact} & 0.16 & Catch Your Crush's Eye & for Girls Only & Boys \\
    \textsc{Interact} & 0.16 & Dance at a School Dance & for Girls & for Guys \\
    \textsc{Present} & 0.21 & Act Like a Kid Again & Girls & Boys \\
    \textsc{Present} & 0.21 & Look Like an Abercrombie Model & for Girls & Boys \\
    \textsc{Present} & 0.21 & Dress Emo & for Girls & Guys \\
    \textsc{Body} & 0.24 & Have Good Hygiene & Girls & Boys \\
    \textsc{Present} & 0.25 & Prepare for a School Dance & for Girls & for Guys \\
    \textsc{Present} & 0.27 & Pack for Soccer Practice & Girls & Boys \\
    \textsc{Interact} & 0.28 & Be in a Female Led Relationship & Women & Men \\
    \textsc{Interact} & 0.30 & Act on a Date & for Girls & for Boys \\
    \textsc{Present} & 0.31 & Dress Cool & for Girls & Guys \\
    \textsc{Body} & 0.31 & Lose Belly Fat & Teen Girls & for Men \\
    \textsc{Present} & 0.33 & Look Hot on Club Penguin & Girls & Guys \\
    \textsc{Present} & 0.35 & Be Awesome & for Girls & for Boys \\
    \textsc{Interact} & 0.36 & Have Fun with Your Friends & Teen Girls & Guys \\
    \textsc{Present} & 0.36 & Dress Like a CEO & Women & Men \\
    \textsc{Interact} & 0.37 & Cradle a Lacrosse Stick & Girls & Men \\
    \textsc{Interact} & 0.41 & Get Your Crush to Like You & Girls & Guys \\
    \textsc{Interact} & 0.43 & Practice Changing Room Etiquette & Girls & Men \\
    \textsc{Interact} & 0.43 & Practice Changing Room Etiquette & Women & Men \\
    \textsc{Body} & 0.44 & Recognize Trichomoniasis Symptoms & Women & Men \\
    \textsc{Present} & 0.45 & Be Popular in Middle School & for Girls & for Boys \\
    \textsc{Body} & 0.47 & Lose Belly Fat & for Women & for Men \\
    \textsc{Body} & 0.49 & Gain Weight Fast & for Women & for Men \\
    \textsc{Body} & 0.52 & Be Indie & for Girls & for Guys \\
    \textsc{Interact} & 0.54 & Grind & for Girls & for Guys \\
    \textsc{Interact} & 0.55 & Host a Sleepover & Teen Girls & for Boys \\
    \textsc{Body} & 0.57 & Treat Acne & Teenage Girls & Teen Boys \\
    \textsc{Body} & 0.60 & Prevent HIV Infection & Women & Men \\
    \textsc{Body} & 0.69 & Recognize Chlamydia Symptoms & for Women & for Men \\

    \midrule
    {} & {} & {} & Indicator for \textbf{K} & Indicator for \textbf{T} \\
    \midrule
    \textsc{Activity} & 0.05 & Flirt & Middle School & for Teens \\
    \textsc{Activity} & 0.10 & Redo Your Bedroom & Preteen Girls & Teen Girls \\
    \textsc{Grown-up} & 0.11 & Look Older & Preteen Girls & Teenage Girls \\
    \textsc{Activity} & 0.14 & Enjoy Summer Vacation & for Kids & for Teens \\
    \textsc{Activity} & 0.18 & Clean Your Room & Kids & Teens \\
    \textsc{Advice} & 0.19 & Enjoy a Plane Ride & for Grade School Kids & Teen Girls \\
    \textsc{Advice} & 0.20 & Be Less Insecure & Preteens & for Teen Girls \\
    \textsc{Activity} & 0.28 & Clean Your Room & Tween Girls & Teens \\
    \textsc{Activity} & 0.29 & Pack for a Vacation & Preteen Girls & Teen Girls \\
    \textsc{Advice} & 0.30 & Get a Boy to Like You & Pre Teens & Teens \\
    \textsc{Activity} & 0.31 & Apply Makeup & Preteens & for Teen Girls \\
    \textsc{Advice} & 0.33 & Balance School and Life & Middle School & Teens \\
    \textsc{Activity} & 0.36 & Host a Girls Only Sleepover & for Preteens & Teens \\
    \textsc{Activity} & 0.39 & Get Ready for Bed & Tween Girls & for Teenage Girls \\
    \textsc{Grown-up} & 0.46 & Get Fit & for Kids & Teenage Girls \\
    \textsc{Activity} & 0.53 & Apply Makeup & Preteens & for Teens \\
    \textsc{Grown-up} & 0.59 & Make Money & for Kids & for Teenagers \\
    \bottomrule
    \end{tabular}
    \caption{All ``Same Title, Different Audience'' guides.} 
    \label{tab:kt_bleu}
\end{table}

\begin{table*}[t]
    \centering
    \begin{tabular}{rclr}
    \toprule
    \textbf{Be~(\ldots)} & $\boldsymbol{X}$ & \textbf{(\ldots)} & \textbf{indicator}\\
    \midrule
     Be & Popular & and Athletic & (for Girls) \\
     Be & Popular & in Grade 6. & (for Girls.) \\
     Be & Popular & in Middle School & (for Girls) \\
     Be & Popular & in a School Uniform & (Girls) \\
     Be & Popular & in Secondary School & (for Girls) \\
     Be a & Cute & Teen & (Girl) \\
     Be & Cute & {} & (Tween Girls) \\
     Be the & Cute & and Hot Teen & (Girls) \\
     Be & Cute & at School & (Girls) \\
     \midrule
     Be & Cool & Around Your Crush & (for Boys) \\
     Be & Cool & in High School & (Boys) \\
     Be a & Cool & Christian & (Teen Guys) \\
     Be & More & Attractive to Girls & (for Boys) \\
     Be & More & Physically Attractive & (Men) \\
     Be & More & Socially Open & (Men) \\
     \midrule
     Be a & Good & Hamster Owner & (for Kids) \\
     Be a & Good & Stuffed Animal Mom & (for Kids) \\
     Be & Good & With Money & (for Kids) \\
     \midrule
     Be a & Good & Friend & (Teens) \\
     Be a & Good & Writer & (Teens) \\
    \bottomrule
    \end{tabular}
    \caption{The most common completions in the titles for ``how to be''. }
    \label{tab:how-to-be-titles}
\end{table*}
\FloatBarrier

\subsection{Classification tasks}\label{sec:appedix-classification}

\begin{table}[htpb]
    \centering
        \begin{tabular}{lr|r}
        \toprule
        \quad \textbf{model-name} & \textbf{W--M} & \textbf{K--T} \\
\texttt{bert-base-uncased}       &  0.57 & 0.64 \\
\texttt{roberta-base}            &  0.81 & 0.73 \\
\texttt{bert-large-uncased}      &  0.73 & 0.74 \\
\texttt{roberta-large}           &  0.82 & 0.75 \\ 
    \bottomrule
    \end{tabular}
    \caption{The performance on the \textsc{dev} set of the classification tasks with optimized LR using the \lbrack CLS\rbrack~token representations from the different LMs.}
    \label{tab:challenge_embeddings}
\end{table}
\FloatBarrier

Most predictive features of the \texttt{combined} model:
\begin{itemize}
    \item[ 
\textbf{W}:] hadn't - wasn't - cute - makeup - ourselves - bag - skirt - outfit - move - sleep
\item[\textbf{M}:] man - product - boy - yourselves - o - dance - theirs - shoe - hers - person
\item[\textbf{K}:] kid - the - adult - name - are - step - were - else - probably - mean
\item[\textbf{T}:] teen - without - than - dress - next - her - want - buy - everyone - ADJ
\end{itemize}

\newpage

\subsection{Confusion Matrices}\label{sec:confusion-m}
\begin{table}[htpb]
    \centering
    \begin{tabular}{lrr|rr}
    \toprule
     {} & {} & \textsc{Dev} & \textsc{Test} & {} \\
     {}& \textbf{W} & \textbf{M} & \textbf{W} & \textbf{M} \\
    \textbf{W} & 0.83 & 0.17 & 0.87 & 0.13 \\
    \textbf{M} & 0.48 & 0.52 & 0.36 & 0.64 \\
    \cmidrule{2-5}
    \textbf{W} & 78 & 16 & 82 & 12 \\
    \textbf{M} & 11 & 12 & 5 & 9 \\
    \bottomrule
    \end{tabular}
    \caption{The confusion matrix for the dev set (left) and the confusion matrix for the test set (right).}
    \label{tab:cm_wm}
\end{table}
\begin{table}[!h]
    \centering
    \begin{tabular}{lrr|rr}
    \toprule
    {} & {} & \textsc{Dev} & \textsc{Test} & {} \\
     & \textbf{K} & \textbf{T} & \textbf{K} & \textbf{T} \\
    \textbf{K} & 0.78 & 0.22 & 0.87 & 0.13 \\
    \textbf{T} & 0.35 & 0.65 & 0.30 & 0.70 \\
    \cmidrule{2-5}
     & \textbf{K} & \textbf{T} & \textbf{K} & \textbf{T} \\
    \textbf{K} & 35 & 10 & 33 & 5 \\
    \textbf{T} & 13 & 24 & 11 & 26 \\
    \bottomrule
    \end{tabular}
    \caption{The confusion matrix for the dev set (left) and the confusion matrix for the test set (right).}
    \label{tab:cm_kt}
\end{table}

\end{document}